%% file: 0_acl_latex.tex
\pdfoutput=1

\documentclass[11pt]{article}

\usepackage[]{acl}

\usepackage{times}
\usepackage{latexsym}

\usepackage[T1]{fontenc}

\usepackage[utf8]{inputenc}

\usepackage{microtype}

\usepackage[linesnumbered, ruled]{algorithm2e}
\usepackage[T1]{fontenc}
\usepackage{dsfont}

\usepackage{mdframed}
\usepackage{booktabs}
\usepackage{graphicx}
\usepackage{subcaption}

\usepackage[normalem]{ulem}
\usepackage{enumitem}
\usepackage{xcolor}
\usepackage{soul}
\colorlet{soulred}{red!30}
\sethlcolor{soulred}%

\usepackage{amsmath,amsthm,amsfonts,amssymb,bm}
\DeclareMathOperator*{\argmax}{arg\,max} 
\usepackage{framed}
\usepackage{varioref}
\usepackage{float}
\usepackage{multirow}
\usepackage{dashrule}
\usepackage{url}
\usepackage{pifont}
\newcommand{\eat}[1]{\ignorespaces}

\usepackage{xparse}
\NewDocumentCommand{\heng}
{ mO{} }{\textcolor{red}{\textsuperscript{\textit{Heng}}\textsf{\textbf{\small[#1]}}}}

\title{Retrieval-Augmented Generative Question Answering \\ for Event Argument Extraction}

\author{Xinya Du \\
  Department of Computer Science \\
  The University of Texas at Dallas \\
  \texttt{xinya.du@utdallas.edu} \\\And
  Heng Ji \\
  Department of Computer Science \\
  University of Illinois Urbana-Champaign \\
   \texttt{hengji@illinois.edu}
   }

\begin{document}
\maketitle

\begin{abstract}
Event argument extraction has long been studied as a sequential prediction problem with extractive-based methods, tackling each argument in isolation. Although recent work proposes generation-based methods to capture cross-argument dependency, they 
require generating and post-processing a complicated target sequence (template). 
Motivated by these observations and recent pretrained language models’ capabilities of learning from demonstrations. We propose a retrieval-augmented generative QA model (R-GQA) for event argument extraction. It retrieves the most similar QA pair and augments it as prompt to the current example's context, then decodes the arguments as answers. Our approach outperforms substantially prior methods across various settings (i.e. fully supervised, domain transfer, and few-shot learning). 
%
Finally, we propose a clustering-based sampling strategy (JointEnc) and conduct a thorough analysis of how different strategies influence the few-shot learning performance.\footnote{The implementations will be released at \url{https://github.com/xinyadu/RGQA}.}
\end{abstract}

\input{1_intro}

\input{3_method}

\input{4_experiment_analysis}

\input{5_related}

\input{7_conclusion}

\input{6_ack.tex}

\bibliography{custom}
\bibstyle{acl_natbib}

\appendix
\input{appendix/appendix}

\end{document}

%% file: 1_intro.tex
\begin{figure}[ht]
\centering
\resizebox{\columnwidth}{!}{
\includegraphics{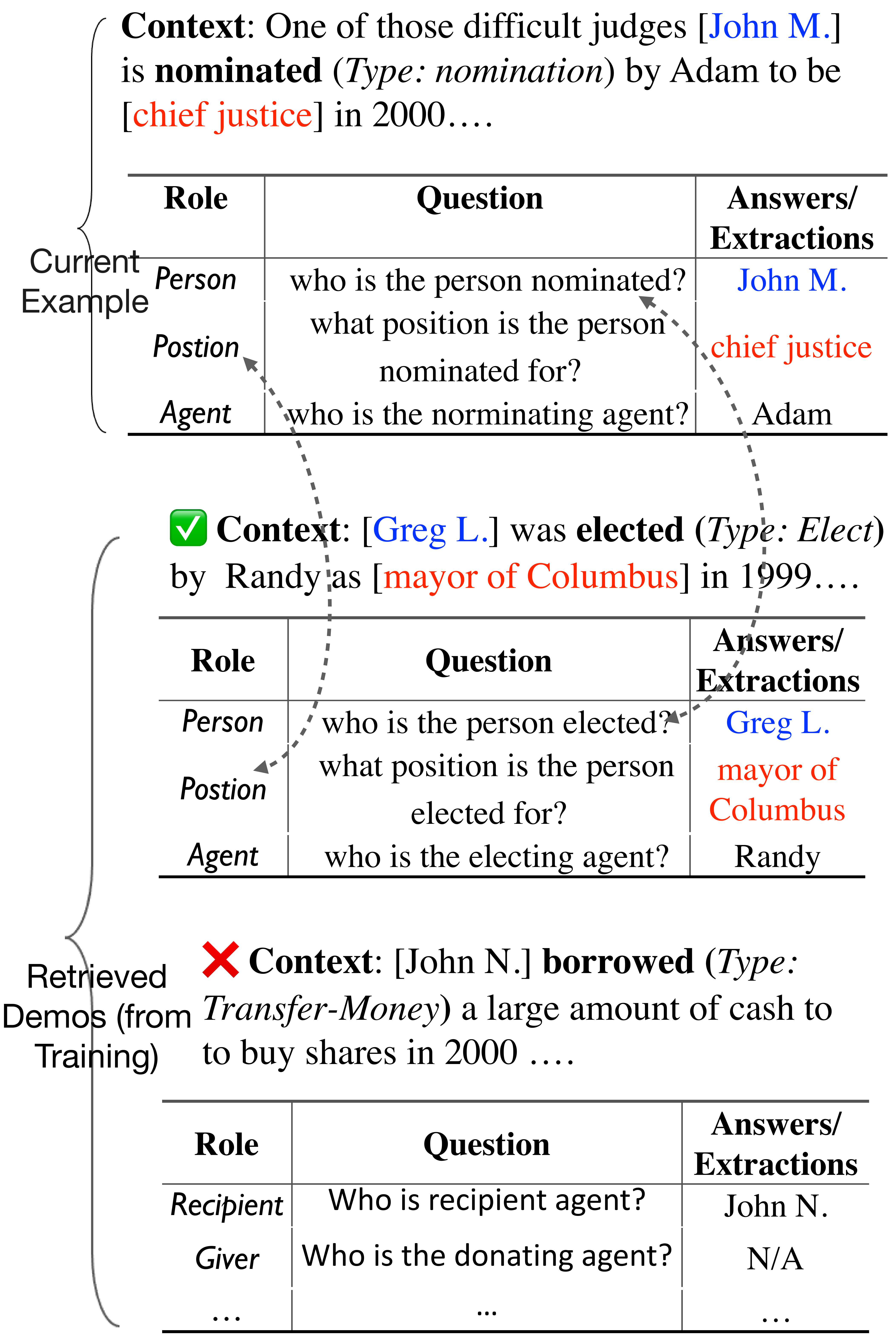}
}
\caption{Current/test example's context and question for each role have great similarities to the retrieved demonstrations (context and QA pairs).}
\label{fig:eg}
\end{figure}

\section{Introduction}

Many documents report sequences of events corresponding to common situations in the real world. Arguments of different roles provide fine-grained understanding of the event (e.g. {\sc individuals}, {\sc organizations}, {\sc locations}) and also influence the determination of the event type~\cite{grishman2019twenty}.
%
As compared to detecting the trigger (usually verbs) of an event, extracting arguments involve recognizing mention spans (consisting of multiple words) of various roles across sentences~\cite{jurafsky2018speech}. We list an example in Figure~\ref{fig:eg}, given the context and the event type ({\it nomination}), all arguments for the three roles (i.e {\sc Person}, {\sc Position}, {\sc Agent}) should be extracted.

To overcome the error propagation of extractive models \cite{li-etal-2013-joint, du-cardie-2020-event} and efficiently capture the cross-role dependencies, end-to-end template generation-based information extraction approaches~\cite{li-etal-2021-document, huang-etal-2021-document, du-etal-2021-template} have been proposed.
However, they 
(1) suffer from the dense output template format (fewer training instances) and cannot fully exploit semantic relations between roles with the constrained templates;
(2) are unable to unleash the excellent analogical capability of large pre-trained models~\cite{brown2020language} on similar input-output pairs to produce extraction results.

Based on our observations in the real circumstances, examples often bear great similarities (in terms of both syntax and semantics) with other examples (Figure~\ref{fig:eg}). In this Figure, we have current input context ``... difficult judges John M. is nominated ...'' for a nomination event. 
When searching through examples in the large store (e.g. training set) for {\it demonstrations (input-output pairs}\footnote{In our QA setting, input consists of the context and question (for each argument role), output consists of the arguments.}), the two most similar examples' input-output pairs are presented. 
Both of the retrieved examples' contexts have large semantic similarities with the context of the current example.
The first retrieved example's questions (for each role) also match the input examples'. The second example's questions do not.
Thus, to help the model determine ``how much'' to learn from the demonstrations is also important. 

Motivated by the weaknesses of previous methods and our observations, we introduce a {\it retrieval-augmented generative question answering} model ({\bf R-GQA}) for event argument extraction. Firstly, our formulation for event extraction as a generative question answering task enables the model to take advantage of both question answering (exploiting label semantics) and text generation, and there's no need for threshold tuning. We conduct experiments on two settings (1) fully-supervised setting\footnote{train and test both on ACE05~\cite{doddington-etal-2004-automatic}.} and (2) domain transfer setting\footnote{train on ACE05 and test on WikiEvent~\cite{li-etal-2021-document}.}. Empirically, our method outperforms previous methods (extraction QA and template generation-based methods) substantially ({\bf Contribution 1}).

To enable our generative model based on large pretrained model to explicitly learn (``reason'') from similar demonstrations as prompt, we add to our model a retrieval component. It uses similarity/analogy score to decide how much to rely on retrieved demonstrations. It significantly outperforms the generative QA model (our proposed baseline without the retrieval component) in both settings ({\bf Contribution 2}).
%
%
%
What's more, we also investigate various models' performance in the few-shot extraction setting. 
As far as we know, there's a large variance in terms of performance when the examples for training/evaluation are randomly sampled, causing different methods not comparable.
Thus 
(1) we investigate models' behavior in the few-shot event extraction setting on different sampling strategies (e.g. random, clustering-based) and how the model performance and distribution distance (between true data and sampled data) correspond;
(2) we design a clustering-based sampling strategy (JointEnc),
which selects the most representative (unlabeled) examples by leveraging both context \& trigger embedding. It is better than random sampling and one-round active learning. Our discussions on sampling methods help improve benchmarking models' few-shot setting performance ({\bf Contribution 3}).

%% file: 3_method.tex
\section{Problem and Definitions}
\label{sec:def}
\paragraph{Event Ontology, Templates, and Questions}
We focus on extracting event arguments from a sequence of words. An event consists of (1) a trigger and the type ($E$) of the event; (2) corresponding arguments $\{arg^{E}_{1}, arg^{E}_{2}, ...\}$ for event type $E$. Both the event type and argument roles are pre-defined in the ontology. 
%
Apart from the event types and argument roles, the ontology also provides definitions and templates for the argument roles. For example, when $E = \textit{Movement-Transportation-Evacuation}$, 
the template for the argument roles is provided,
\begin{equation}
\begin{gathered}
\nonumber
[arg_1] \ \ \text{transported} \ \ [arg_2] \ \ \text{in} \ \ [arg_3] \\ 
\text{from} \ \ [arg_4] \ \ \text{place to} \ \ [arg_5] \ \ \text{place.}
\end{gathered}
\end{equation}
Based on the definitions of argument roles and the templates in the ontology, we can generate the natural questions for each argument role based on the mechanism proposed in \newcite{du-cardie-2020-event}.
For example, in this example, $arg_1$ (\textsc{Transporter}):``who is responsible for transport'', $arg_2$ (\textsc{Passenger}):``who is being transported'', $arg_3$ (\textsc{Vehicle}):``what is the vehicle used'', $arg_4$ (\textsc{Origin}):``where the transporting originated'', $arg_5$ (\textsc{Destination}):``where the transporting is directed''\footnote{For the full list of questions for WikiEvent argument roles, please refer to the Appendix Sec E.}.

\paragraph{Demonstrations Store}
\newcite{brown2020language} proposed to use in-context demonstrations (input-output pairs) as prompt to test the zero-shot performance of large pretrained language models. 
For our retrieval-augmented approach, we denote the set of demonstrations/prompts to choose from $ST$. In this work, we initiate $ST$ with the training set.\footnote{Other external resources can also be added to $ST$.}

\begin{figure*}[t]
\centering
\resizebox{\textwidth}{!}{
\includegraphics{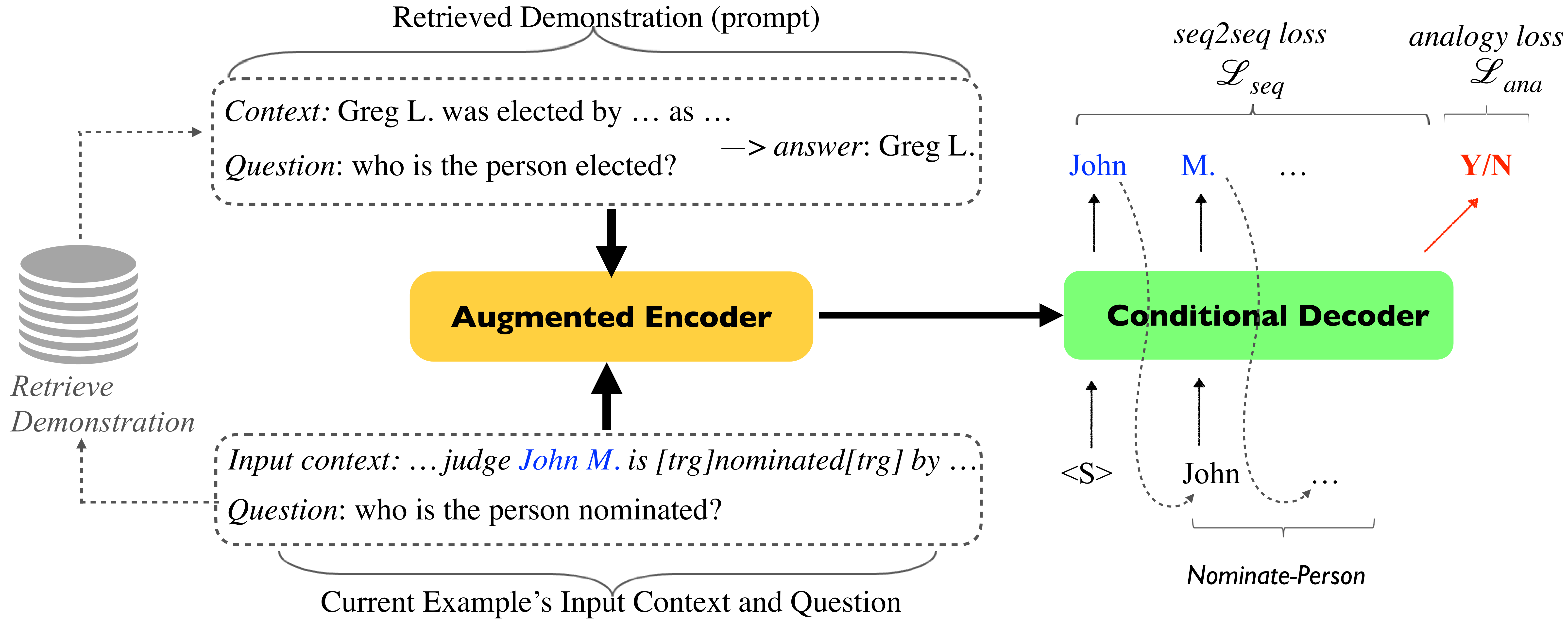}
}
\caption{Our Retrieval-Augmented Generative Question Answering Model.}
\label{fig:framework}
\end{figure*}

\paragraph{Data and Sampling Strategy}
In the fully-supervised setting, we use the entire training set (1) to train the models; (2) as the demonstration store.
In the few-shot setting, motivated by the need to reduce annotation cost, we assume that there is only a fixed budget for annotating $K$ examples' arguments for training, and call the annotated subset $S_{few}$. Then we use $S_{few}$ as both the training set and demonstration store.


\section{Methodology}

We first describe the retrieval-augmented generative question answering model (Figure~\ref{fig:framework}), including (1) the generation model and how to construct the demonstration (prompt) as well as the final input\&target sequence; (2)  training, decoding, post-processing details; and how they differ from template-generation based models. Then we introduce our clustering-based sampling strategy to diversify the training examples for the few-shot setting.


\subsection{Retrieval-Augmented Generative QA}
BART~\cite{lewis-etal-2020-bart} is a large pre-trained encoder-decoder transformer architecture based on \newcite{vaswani2017attention}. Its pretraining objective is to reconstruct the original input sequence (denoising autoencoder). Prior work reports that this objective helps the extraction problems~\cite{li-etal-2021-document, du-etal-2022-dynamic}. Thus we use pre-trained BART as our base model. It is presented in Figure~\ref{fig:framework}. 
For each argument role, 
the R-GQA model's input $\mathbf{x}$ is conditioned on 
(1) the current example's context;
(2) question for the role
and (3) the demonstration store $ST$.
We will explain the details below. 
The ground truth sequence $\mathbf{y}$ is based on the gold-standard argument spans for the current training instance. The goal is to find $\mathbf{\hat{y}}$ such that,
\begin{equation}
\begin{split}
\mathbf{\hat{y}} & = \argmax_{\mathbf{y}} p \left( \mathbf{y} \vert \mathbf{x} \right)  \\ 
\end{split}
\end{equation}
where $p(\mathbf{y}|\mathbf{x})$ is the conditional log-likelihood of the predicted argument sequence $\mathbf{y}$ given input $\mathbf{x}$.

To construct $\mathbf{x}$ and $\mathbf{y}$,
apart from the special tokens in the vocabulary of BART -- including the separation token $[sep]$, and start/end token of a sequence (i.e. $<S>$ and $</S>$), 
we add three new tokens: $[demo]$, $[tgr]$ and $[sep\_arg]$. 
More specifically, $[demo]$ denotes which part of the input sequence is the demonstration/prompt,
$[trg]$ marks the trigger of the event in the input context,
$[sep\_arg]$ is used as the separator token gold arguments.

Given an example (including context and the event trigger), 
for each argument role of the event type $E$, the input format is as follows, where we instantiate all components to obtain the final \textbf{input sequence}:
\begin{equation}
\nonumber
\begin{gathered}
\mathbf{x} \ = \ <S> \ \ [demo] \ \ \text{Demonstration} \ \ [demo] \\ 
\text{Question} \ \ [sep] \ \ \text{Input Context} \ \ </S>
\end{gathered}
\end{equation}
where ``Question'' is from the question set derived from respective ontology (Section~\ref{sec:def});
for ``Input Context'', we mark up the current example's trigger word with $[trg]$ token for emphasizing. For the example in Figure~\ref{fig:framework}, the input context would be ``... John M is $[trg]$ nominated $[trg]$ by ...''.

As for the ``Demonstration'', we first retrieve it from the demonstration store ($ST = \{d_1, d_2, ...\}$) $d_r$ which is most similar to current question and input context, it is a (<Question, Context>, Arguments) pair.
%
We concatenate the components (with the separation tokens in between them) as the final demonstration sequence.
\begin{equation}
\nonumber
\begin{gathered}
\text{Demonstration} \ d_r =  \text{Q}_r \ [sep] \ \text{C}_r \ [sep] \\
 \text{The answer is:} \ \text{A}_r \
\end{gathered}
\end{equation}
We use S-BERT~\cite{reimers-gurevych-2019-sentence} to calculate the similarity scores between the current instance and all demonstrations in $ST$. S-BERT is a modification of the BERT model~\cite{devlin-etal-2019-bert} that uses siamese and triplet network structures to obtain semantically meaningful embeddings for word sequences\footnote{The SentenceTransformer library (\url{https://www.sbert.net/docs/quickstart.html}) supports calculations in batch.}. 

To construct the {\bf target (sequence)}, we first determine how much to learn from the demonstration -- if the similarity score is above a threshold (determined on the development  set), and the demonstration and current instance both have a non-empty answer, then we assign 1 (Yes) to $y_{analogy}$, otherwise 0 (No). 
Then we concatenate all argument spans of the role with $[sep\_arg]$ to construct $\mathbf{y}_{seq2seq}$,
\begin{equation}
\nonumber
\begin{gathered}
\mathbf{y}_{seq2seq} \ = \ <s> \ \text{Argument}_1 \\
\ \ [sep\_arg] \ \ \text{Argument}_2 \ \ [sep\_arg] \ \ ... </s> 
\end{gathered}
\end{equation}
The final $\mathbf{y}$ includes $\mathbf{y}_{seq2seq}$ and $y_{analogy}$.

\subsection{Training and Inference}

\paragraph{Training}

After the preparation for $S = \{(\mathbf{x}^{(i)}, \mathbf{y}^{(i)})\}_{i=1}^{|S|}$, we minimize the joint loss function during training, 
\begin{equation} 
\label{equ:loss}
\begin{split}
\mathcal{L} & =  \mathcal{L}_{seq2seq} + \mathcal{L}_{analogy} \\ 
\mathcal{L}_{seq2seq} & = - \sum_{i=1}^{|S|} \log p (\mathbf{y}_{seq2seq}^{(i)}\vert \mathbf{x}^{(i)}; \theta) \\
& = - \sum_{i=1}^{|S|} \sum_{j=1}^{|\mathbf{y}_{seq2seq}^{(i)}|} \log p (y_j^{(i)}|\mathbf{x}^{(i)}, y_{<j}^{(i)} ; \theta)
\end{split}
\end{equation}
where $\mathcal{L}_{seq2seq}$ is the cross-entropy loss between the decoder’s output and the target sequence $\mathbf{y}_{seq2seq}$.
$\mathcal{L}_{analogy}$ is the binary cross-entropy loss calculated with the final hidden state of the final decoder token.

\paragraph{Inference and Post-processing}
At test time, we conduct greedy decoding to obtain the target sequence, 
then we split the decoded sequence with respect to $[seq\_arg]$. Since it is also required to obtain the offsets of the argument in the input context, we automatically match the candidate argument's span with the input context.
Then, if there's no matched span, we discard the candidate argument; if there are multiple matches, we select the one closest to the trigger word.
For example, if the input context is ``One of those difficult judges [\textit{John M.}] is nominated (Type: nomination) by Adam to be chief justice in 2000.. [John M.] started office on ...'' and there are two appearances of the candidate argument (in brackets) for the role {\sc person}, then we use the first candidate's offsets. Different from our methods, the template-based generation method generates a sequence similar to the one in Section~\ref{sec:def} -- causing the model to (1) not fully exploit the semantic relations of roles across event types; (2) require more complicated post-processing including an additional step to obtain arguments from the generated template.

\subsection{Few-shot Setting and Sampling Strategy}
\label{sec:fewshot_and_sample_strategy}

\begin{algorithm}[h]
\small
\SetKwInOut{Input}{Input}
  \Input{$|S|$ Unlabeled Examples, Sample Size $N$}
  
  $k \leftarrow$ \# event types (based on ontology)\;
  $S_{few} \leftarrow [ \ \ ]$\;
  {\color{blue}\tcp{\small  obtain embeddings for all unlabeled instances}}
  \For{$i \leftarrow 1$ \KwTo $|S|$}
  {
    $rep_i \leftarrow [enc(context_i), enc(trigger\_text_i)]$\;
    add $rep_i$ to $all\_reps$\;
  }
  $clusters = k\_means(all\_reps)$\;
  {\color{blue}\tcp{\small  add instances to samples}}
  \For{$i \leftarrow 1$ \KwTo $k$}
  {
    $\#instance = \frac{length(clusters[i])}{|S|}*N$\;
    $instances = sample(clusters[i], \#instances)$\;
    add $instances$ to $S_{few}$\;
  }
\caption{\small Our Strategy for Obtaining $S_{few}$}
\label{algo:cluster}
\end{algorithm}

In the few-shot setting, we assume that we have a budget to obtain annotations for a limited number of examples' arguments (5\%-20\% of all examples) for training.
We denote the set of few training examples as $S_{few}$. We study 
(1) how different sampling strategies affect the $S_{few}$'s distributions and models' performance; 
(2) how to select the best set of examples (in zero or one round\footnote{One-round active learning setting~\cite{wang2021oneround}.}) and have them annotated for training, to achieve better performance at test time.

We propose a sampling method called {\bf JointEnc}. It uses k-means clustering upon the embeddings of both {\it input context} and {\it trigger text}. This is easier to implement as compared to the one-round active learning setting since our method does not require iterative training/testing for selecting unlabeled examples. Details of how we obtain $S_{few}$ are illustrated in Algorithm~\ref{algo:cluster}.
Specifically, we first obtain embeddings of context and trigger text for each unlabeled example (line 3-6). Then we conduct k\_means based clustering upon the embeddings (line 7). Finally, we calculate the proportions of examples across all clusters\footnote{We also try adding average number of examples for each cluster but performance is substantially worse.}; and add the corresponding number of examples of each cluster to $S_{few}$ (line 8-12).

%% file: 4_experiment_analysis.tex
\section{Experiments and Analysis}

We conduct experiments and compare our model to baselines in three settings on two datasets: 
(1) full supervision setting; domain transfer setting; as well as (3) few-shot training setting (Section~\ref{sec:fewshot}).

\input{tables/data_stats}

\subsection{Datasets Statistics and Evaluation}
For the fully-supervised experiments, we use ACE 2005 corpus for evaluation, it contains documents crawled between year 2003 and 2005 from a variety of areas. We use the same data split and preprocessing steps as in previous work~\cite{wadden-etal-2019-entity, du-cardie-2020-event}.
For the domain transfer setting, we conduct training on the ACE05 training set and test on the WikiEvent test set. WikiEvent contains real-world news articles annotated with the DARPA KAIROS ontology\footnote{\url{https://www.darpa.mil/news-events/2019-01-04}}. Most of the event/argument types of WikiEvent's ontology do not appear in the ontology of ACE05 (e.g. Disaster, Cognitive, Disease).

The statistics of the datasets are shown in Table~\ref{tab:dataset}. We use the same test set as in \newcite{li-etal-2021-document} in the domain transfer setting. 
As for the preprocessing step of WikiEvent, since we train the models on the ACE05 (including only arguments in the sentence where each trigger appears), we also use arguments within a maximum context window of the length equal to the average of ACE05 sentence length).

\input{tables/tab_full}

As for the evaluation, we use the same criteria as in previous work~\cite{li-etal-2013-joint} to judge whether an extracted argument is correct. We consider an argument mention to be correctly identified if its offsets match any of the reference arguments of the current event (i.e. argument identification, or Arg Id. for short); and an argument is correctly classified if its role also matches (i.e. argument classification or Arg C.).

When comparing the extracted argument spans with the gold-standard ones, in addition to using extract match (EM), we also consider head noun phrase match (HM). It is more lenient than EM since it does not require the boundary/offsets to be matched correctly~\cite{huang2012modeling, du-cardie-2020-document}. 
For example, ``the John M.” and “John M.'' match under the HM metric.
Our results are reported with Precision (P), Recall (R), and F-measure (F1) scores.
%


\subsection{Baselines}
We compare our model to several representative and competitive baselines (extractive methods and generation-based methods). 
{\bf EEQA}~\cite{du-cardie-2020-event} uses the pretrained BERT as the base model and add a linear layer on top, to obtain the beginning and end offsets of the answer/argument spans in the input context for each role.
%
{\bf GenIE}~\cite{li-etal-2021-document} use template-based generation for argument extraction. Its objective is to generate the template (including the arguments) and post-process the generated template to obtain the argument mentions~(Section~\ref{sec:def}). Sometimes the generated sequences don't conform to the original template thus affecting the performance.
{\bf Generative QA} is our own baseline without the retrieval component -- it directly encodes the question for the current argument role and input context to generate the candidate argument spans.

\subsection{Fully-Supervised Setting Results}

In Table~\ref{tab:full}, we report results for the fully supervised setting. 
The score for Argument identification is strictly higher than Arg. classification since it only requires both the mention span match and role match.
%
We denote our proposed framework as R-GQA. To find out how the explicit modeling of the analogical relations (semantic relatedness) between the demonstration and the current instance helps, we also report ablation study results. More specifically, we use \verb|BART-Large| for all methods that use BART as the base model to ensure they are comparable. For our own model and its variations, we conduct three runs, and calculate the average of their performance and standard deviations.

\input{tables/domain_transfer}

\input{tables/fewshot_compare}

We observe that:
(1) all the text generation-based approaches outperform substantially EEQA (the extractive question answering based approach) in both precision\&recall; 
Plus, generation-based methods require only one pass and are faster than extractive-based method which has $O(n^2)$ complexity for span enumeration;
(2) Our methods based on generative QA (with 17621 gold QA pairs) substantially improve over the pure template-generation based method (with 4419 gold templates), we see that the better F1 mainly comes from consistently increase of precision\&recall (\textasciitilde3\%-4\% for EM, \textasciitilde1.5\%-2\% for HM). It makes sense considering in the template generation setting (I) hallucination happens; and (II) the generation sequence is longer, as compared to generating arguments for only one role in one pass;
(3) Our R-GQA method benefits greatly from the retrieved demonstrations (prompts). We see that the better performance mainly comes from the increase in recall (smaller variance).
Moreover, as for the functionality of explicitly model analogy relation ($\mathcal{L}_{analogy}$), we find that it provides a boost of recall of around 3\% without sacrificing precision. These to a certain extent prove that the demo's QA pair encourages the model to generate more arguments for the current instance.

\subsection{How Does R-GQA perform in the domain transfer setting}
\label{sec:transfer}

To mimic the real-world setting, we examine the portability of the models to test set of a new ontology (event types and argument types) such as in \newcite{li-etal-2021-document}. More specifically, we conduct training on ACE05 (with 33 event types) and test on WikiEvent dataset (with 50 event types). 

In Table~\ref{tab:domain_trans}, we present the domain transfer results.
For this new setting, the best methods' performance on WikiEvent are around 20\% lower (F1)
as compared to the fully supervised setting~\cite{du-etal-2022-dynamic}.
Mainly because:
(1) the WikiEvent dataset is harder as compared to ACE05 -- with a performance drop around 5-10\% F1 across models;
(2) the test set of WikiEvent includes many event/argument types that are distinct from existing ones from ACE05. Accordingly, we find that performance on the subset of data of distinctly event types largely drops. We list the types in Appendix~\ref{sec:distinct}.
When comparing QA-based generation model and GenIE, we observe that 
(1) recall of the QA-based models is substantially higher (>10\%) -- leading to large argument identification performance improvement; while our models do not have an advantage in precision and even drops a bit, but the general performance (F1) is consistently higher;
(2) Our R-GQA model's retrieval component helps the model generate more arguments and improves R and F1.

\subsection{How Does R-GQA perform in Few-shot Setting and What is Sampling Strategy's Influence}
\label{sec:fewshot}

Firstly, in Table~\ref{tab:fewshot_compare}, we present comparisons between GenIE and R-GQA in the few-shot setting on ACE05. To obtain the few-shot training examples, we use the sampling strategy proposed in Section~\ref{sec:fewshot_and_sample_strategy}.
The \# examples varies from 200 (5\%) to 1k (20\%). We observe the trend that when the number of examples is smallest, the performance gap is largest (around 10\% F1). While as the example number grows, generally the gap minimizes -- from 10\% (200), to 6\% (600), to around 2\% (1k).



\input{figures/fig_sampling_strategy}

Next, we report results for different sampling methods (including the one-round active learning setting) to find out what are the more important factors for the event argument extraction task's annotation (with a fixed budget).
Namely, we sample from ``unlabeled'' examples with the following strategies:
{\bf Random} picks the examples randomly which (nearly) match the distribution of event types in the test set;
{\bf AL} is the one-round active learning based approach -- basically, a model is trained on the 100 examples with annotations and unlabeled examples that are most challenging (model most uncertain about) are selected.
Our {\bf JointEnc} strategy first conducts clustering on unlabeled examples (based on {\it both input context and trigger text}) and selects from each cluster \# examples proportional to the size of each cluster;
{\bf Context} also conducts clustering based sampling similar to JointEnc but only embeds each example based on its context.

For the few-shot setting with increasing sampling size, we calculate the Hellinger distance~\cite{beran1977minimum} between distributions of examples sampled from each strategy and the true data distribution (represented by training data with labels). The distances are presented in Figure~\ref{fig:dist_gap}. We observe that (1) the distances between distributions of sampled examples and true data distribution decrease, as the sampling size grows; (2) sampled data based on JointEnc is generally closest to true data distribution across different sampling sizes.
Correspondingly, Figure~\ref{fig:fewshot_perf} reports the performances of R-GQA trained on samples from each strategy. The model trained on examples from our JointEnc outperforms other strategies', demonstrating the benefit of JointEnc. 
Moreover, we find that there is a correlation between distribution distances and few-shot experimental results -- the smaller the distances are, models trained on the sampled set have better performance. This phenomenon is especially obvious when the sampling size is small (5\%--10\% of training data).
We also provide an analysis of each event type in Appendix (Section~\ref{sec:distribution_distance_event_type}).

%% file: tables/data_stats.tex
\begin{table}[t]
\centering
\resizebox{\columnwidth}{!}{
\begin{tabular}{lcccccc}
\toprule
& \multicolumn{3}{c}{\textbf{ACE05}} & \multicolumn{3}{c}{\textbf{WikiEvent}} \\ 
 & Train    & Dev      & Test & Train    & Dev      & Test    \\ \midrule
\# event types              & 33                         & 22                       & 31                        & 49       & 35       & 34      \\
\# arg. roles           & 22                         & 22                       & 21                        & 57       & 32       & 44      \\
\# docs                     & 529                        & 40                       & 30                        & 206      & 20       & 20      \\
\# sentences                & 17172                      & 923                      & 832                       & 5262     & 378      & 492     \\
\begin{tabular}[c]{@{}l@{}}avg \# events \\ per doc\end{tabular} & 9.26                       & 16.71                    & 10.58                     & 15.73    & 17.25    & 18.25 \\ \bottomrule 
\end{tabular}}
\caption{Dataset Statistics.}
\label{tab:dataset}
\end{table}

%% file: tables/tab_full.tex
\begin{table*}[t]
\centering
\resizebox{\textwidth}{!}{
\begin{tabular}{l|ccc|ccc}
\toprule
\multicolumn{1}{c|}{\multirow{2}{*}{EM}} & \multicolumn{3}{c|}{Arg Identification}               & \multicolumn{3}{c}{Arg Classification}               \\
                   & P                & R               & F1              & P                & R               & F1              \\ \midrule
\begin{tabular}[c]{@{}l@{}}EEQA \\ \cite{du-cardie-2020-event} \end{tabular} & 69.16            & 62.65           & 65.74           & 66.51            & 60.47           & 63.34           \\
GenIE~\cite{li-etal-2021-document}  & 71.13            & 68.75           & 69.92           & 67.82            & 65.55           & 66.67           \\
Generative QA            & 75.40 $\pm$ .70  & 72.10 $\pm$ .26 & 73.71 $\pm$ .20 & 71.92 $\pm$ .88  & 69.09 $\pm$ .59 & 70.47 $\pm$ .12 \\
R-GQA             & 76.90 $\pm$ 1.04 & 74.17 $\pm$ .73 & 75.51 $\pm$ .58 & 74.10 $\pm$ .97  & 71.46 $\pm$ .47 & 72.75 $\pm$ .36 \\ \midrule
\multicolumn{7}{c}{Ablations} \\
\quad w/o analogy loss & 76.20 $\pm$ 1.27 & 72.04 $\pm$ .97 & 74.06 $\pm$ .33 & 73.90 $\pm$ 1.39 & 69.87 $\pm$ .73 & 71.82 $\pm$ .32 \\ \bottomrule
\end{tabular}
}

\smallskip
\smallskip
\bigskip

\resizebox{\textwidth}{!}{
\begin{tabular}{l|ccc|ccc}
\toprule
\multicolumn{1}{c|}{\multirow{2}{*}{HM}} & \multicolumn{3}{c|}{Arg Identification}               & \multicolumn{3}{c}{Arg Classification}               \\
                   & P                & R               & F1              & P                & R               & F1              \\ \midrule
GenIE~\cite{li-etal-2021-document}  & 72.85 & 69.12 & 70.94 & 69.92 & 66.50 & 68.17   \\
Generative QA          &  75.45 $\pm$ .58  & 73.70 $\pm$ .21 & 74.56 $\pm$ .18 & 71.88 $\pm$ .76  & 70.20 $\pm$ .00 & 71.03 $\pm$ .37 \\
R-GQA           &  76.95 $\pm$ 1.34 & 74.93 $\pm$ .52 & 75.93 $\pm$ .91 & 74.04 $\pm$ 1.00 & 72.10 $\pm$ .21 & 73.05 $\pm$ .59 \\ \midrule
\multicolumn{7}{c}{Ablations} \\
\quad w/o analogy loss & 77.04 $\pm$ 1.32 & 71.88 $\pm$ .52 & 74.36 $\pm$ .34 & 74.86 $\pm$ 1.26 & 69.84 $\pm$ .51 & 72.26 $\pm$ .31 \\ \bottomrule
\end{tabular}
}
\caption{Fully-supervised setting experimental results (in \%) on ACE05 data. The upper table is based on Exact Match (EM) and the bottom table is based on Head Head (HM).}
\label{tab:full}
\end{table*}

%% file: tables/domain_transfer.tex
\begin{table*}[t]
\centering
\small
\begin{tabular}{l|ccc|ccc|ccc|ccc}
\toprule
& \multicolumn{6}{c|}{EM}  & \multicolumn{6}{c}{HM} \\ \cline{2-13}
 \multirow{2}{*}{Models} & \multicolumn{3}{c|}{Arg Id.} & \multicolumn{3}{c|}{Arg C.} & \multicolumn{3}{c|}{Arg Id.} & \multicolumn{3}{c}{Arg C.} \\
& P       & R       & F1     & P       & R       & F1     & P       & R       & F1     & P       & R       & F1     \\ \midrule
\begin{tabular}[c]{@{}l@{}}GenIE \\ \cite{li-etal-2021-document} \end{tabular} & 49.96 & 23.47 & 31.88 & 44.92 & 21.09 & 28.66 & 52.87 & 24.84 & 33.74 & 46.94 & 22.04 & 29.95 \\
Generative QA          & 47.12   & 35.61   & 40.57  & 32.32   & 24.42   & 27.82  & 49.71   & 37.57   & 42.79  & 34.20   & 25.84   & 29.44  \\
R-GQA & 44.88   & 40.68   & {\bf 42.63}  & 31.42   & 28.42   & {\bf 29.82}  & 47.65   & 43.17   & {\bf 45.25}  & 33.10   & 29.93   & {\bf 31.41} \\ \bottomrule
\end{tabular}
\caption{Domain transfer setting results (in \%).}
\vspace{-0.2cm}
\label{tab:domain_trans}
\end{table*}

%% file: tables/fewshot_compare.tex
\begin{table*}[t]
\centering
\small
\resizebox{\textwidth}{!}{
\begin{tabular}{l|ccccccccc}
\toprule
& 200   & 300   & 400   & 500    & 600    & 700    & 800    & 900    & 1000   \\
Models & (4.8\%) & (7.1\%) & (9.5\%) & (11.9\%) & (14.3\%) & (16.7\%) & (19.0\%) & (21.4\%) & (23.8\%)  \\ \midrule
GenIE        & 29.13 & 38.19 & 44.19 & 49.09 & 50.26 & 46.85 & 54.41 & 58.47 & 59.94 \\
Ours (R-GQA)  & 38.79 & 47.64 & 52.55 & 56.97 & 56.40 & 58.90 & 61.24 & 58.77 & 61.41 \\
\bottomrule
\end{tabular}
}
\caption{Few-shot performance comparison (F1 in \%).}
\vspace{-0.2cm}
\label{tab:fewshot_compare}
\end{table*}

%% file: figures/fig_sampling_strategy.tex
\begin{figure}[ht]
\centering
\resizebox{\columnwidth}{!}{
\includegraphics{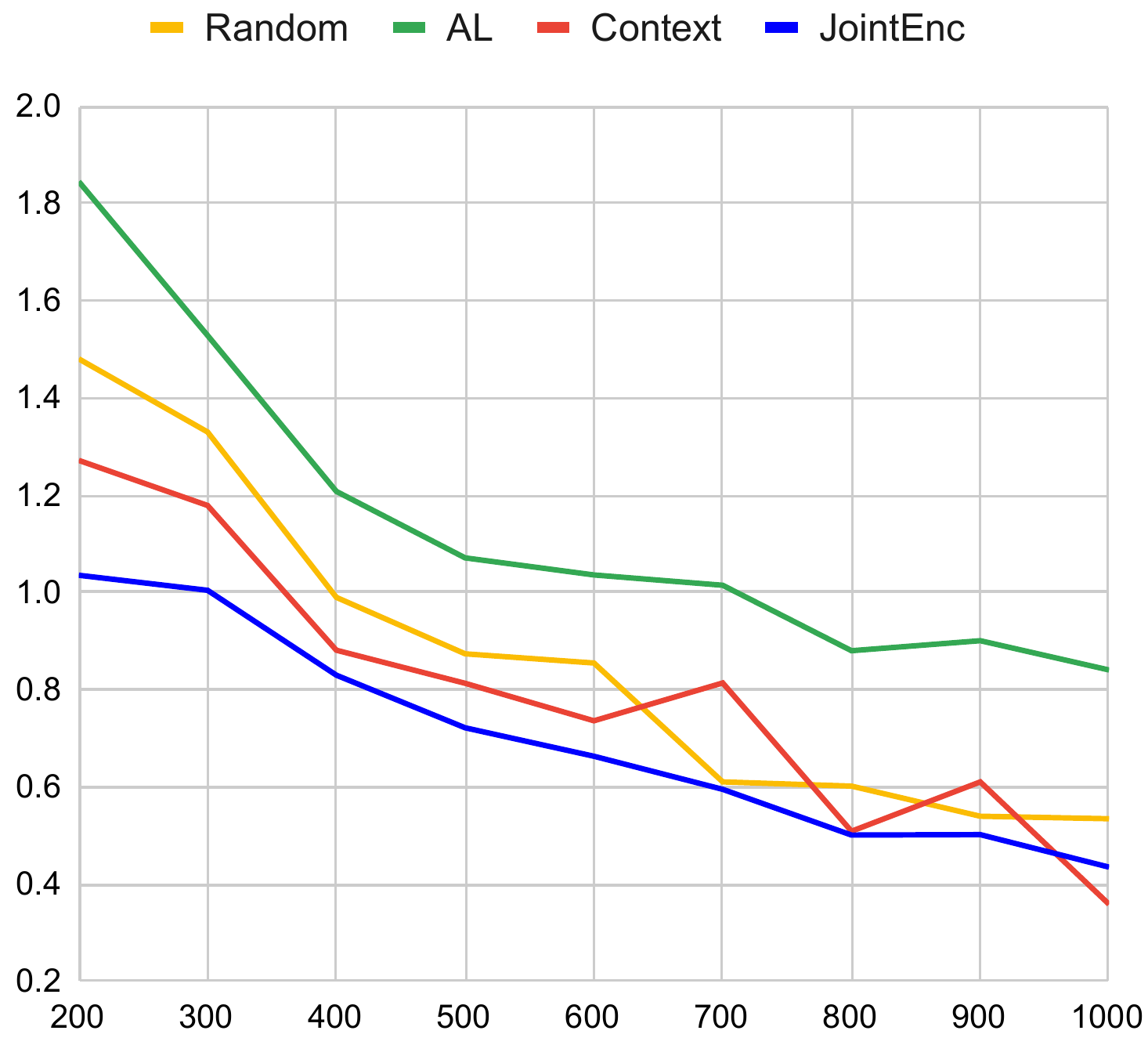}
}
\caption{Distance (Y-axis) between event type distributions of (1) sampled examples with different sampling strategies and (2) real data. X-axis: sampling size.}
\label{fig:dist_gap}
\end{figure}

\begin{figure}[ht]
\centering
\resizebox{\columnwidth}{!}{
\includegraphics{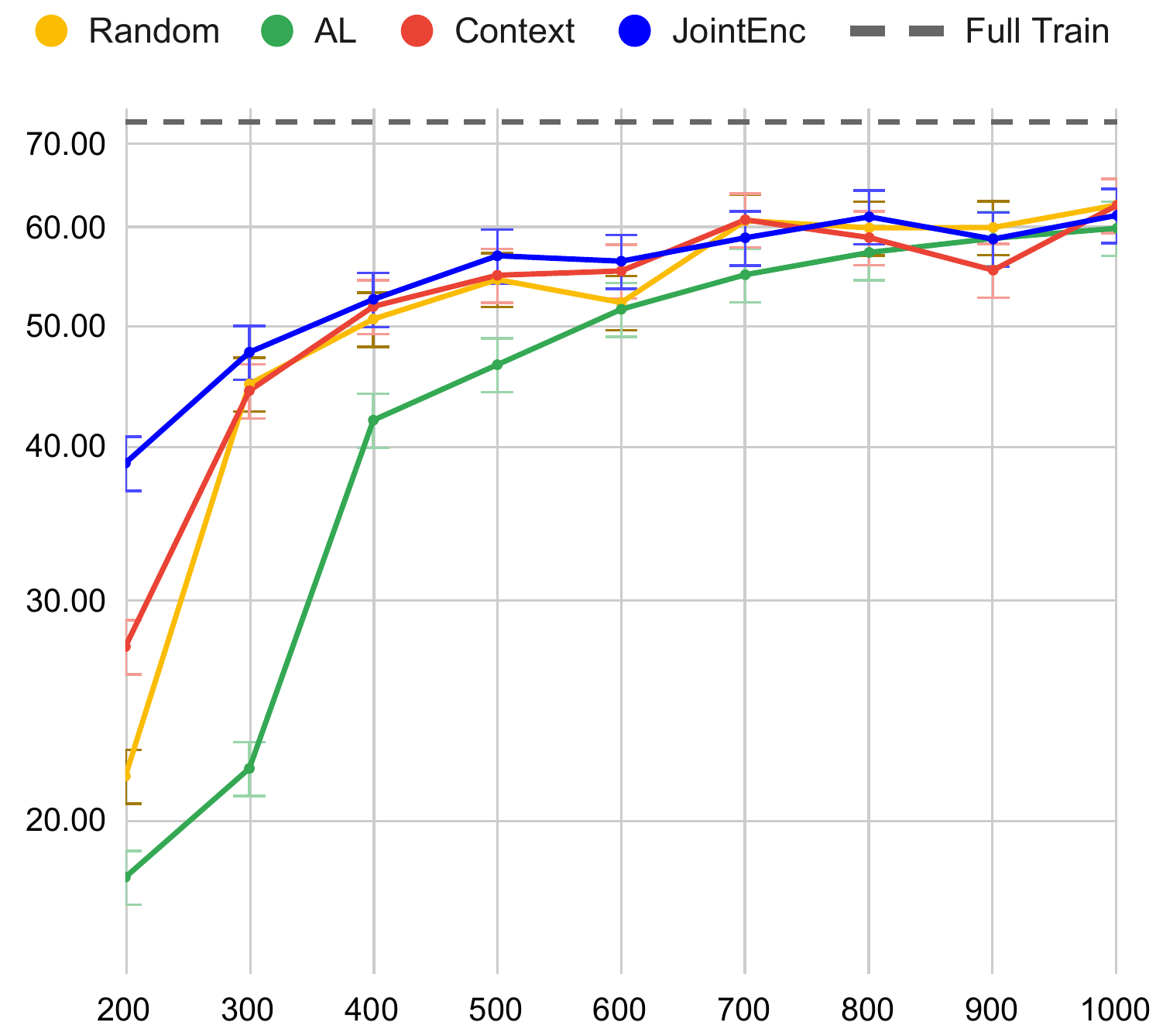}
}
\caption{R-GQA's few-shot performance under different sampling strategies.}
\label{fig:fewshot_perf}
\end{figure}

%% file: 5_related.tex
\section{Related Work}
\paragraph{Event Extraction and Extractive\&Generation-based Approaches}
Traditionally, researchers have been investigating extractive approaches for event/information extraction. Specifically, one branch of work use B-I-O sequence labeling based approaches using CRF or structured SVM models~\cite{bjorne2009extracting, yang-mitchell-2016-joint, lin-etal-2020-joint}, and more recently with neural networks~\cite{chen-etal-2015-event, nguyen-etal-2016-joint-event}. Another branch of extractive approaches includes using span enumeration~\cite{wadden-etal-2019-entity}, as well as using question answering to encourage transfer between argument roles~\cite{du-cardie-2020-event}.

Recently, generation-based approaches have been proposed. 
Among them more generally, TANL~\cite{paolini2020structured} proposes to use translation-based approaches for structured prediction. More specifically, it constructs decoding targets by inserting text markers and labels around entity mentions in the input sentence.
To better capture cross-entity dependencies. \newcite{huang-etal-2021-document, li-etal-2021-document, du-etal-2021-template, huang2022multilingual} propose template-generation based approaches. They fill in the role slots in the template (e.g. Sec \ref{sec:def}) with arguments to construct the gold sequences.
As compared to TANL and template generation-based methods, our R-GQA is designed to be a QA-based generative model with a simpler generation objective. Plus, it augments the current example's context with the most similar demonstration in the training set as prompt.
It gets the best of both worlds (i.e. question answering and generative models).

\paragraph{Retrieval-augmented Text Generation and In Context Learning}

Recent studies have shown the effectiveness of {\it retrieval augmentation} in many generative NLP tasks, such as knowledge-intensive question answering~\cite{lewis2020retrieval, guu2020realm} and dialogue response generation~\cite{cai2019retrieval}. They mainly retrieve additional knowledge or relevant information, but not demonstrations (input-output pairs).
%
Another closely relevant branch of work is {\it in-context learning}, it's a tuning-free approach that adapts to a new task by providing demonstrations (input-output pairs) as prompts to generate the ``answer''~\cite{brown2020language}.
GPT-3 proposes to use random examples as demonstrations. 
\newcite{liu2022makes} refines the strategy by proposing to retrieve demonstrations that are semantically-similar to the current example as prompt. They show the capability of PLM to learn from similar examples.

Different from the work above, our work draws insights from both retrieval-augmented text generation and in-context learning. It (1) retrieves from the training set the most similar demonstration (QA pair) and uses it as a prompt; (2) uses gradient descent to optimize the model.
Plus, it focuses on the specific argument extraction problem -- our model not only augments the input context with demonstration but also determines how much to learn from it (by training with analogical loss).

%% file: 7_conclusion.tex
\section{Conclusions}
In this work, we introduce a retrieval-augmented generative question answering framework (R-GQA) for event argument extraction. 
Our model generates arguments (answers) for each role, conditioned on both the current input context and the analogical demonstration prompt (based on their semantic similarity).
Empirically, we show that R-GQA outperforms current competitive baselines with large margins in fully-supervised, cross-domain and few-shot learning settings.
We conduct a thorough analysis and benchmark how different sampling strategies influence models' performance in the few-shot learning setting.
We find that for event argument extraction, {\it diversifying the examples} makes the sampling distribution closer to the true distribution and contributes to models' better performance.

\section*{Limitations}




This work has certain limitations.
\begin{itemize}
    \item Firstly, since the pre-trained model we use (\verb|BART-Large|) has many parameters, one model's training will nearly occupy one NVIDIA Tesla V100 16GB GPU; As for inference, it takes about 1GB of space.
    \item Although the BART-based models (GenIE and R-GQA) are end-to-end and have a great performance boost, the inference time (about 2 examples/s) is slightly longer as compared to manual-feature based approaches.
    \item In the real domain transfer setting, the general performance of models is still lower than 40\% (F1), making the systems not competitive in real circumstances. In the future, it is worth investigating how to tackle this challenge by both more general ontology designing and stronger\&robust methods.
\end{itemize}

%% file: 6_ack.tex
\section*{Acknowledgement}
We thank the anonymous reviewers helpful suggestions. 
This research is based upon work supported by U.S. DARPA KAIROS Program No. FA8750-19-2-1004, U.S. DARPA AIDA Program No. FA8750-18-2-0014 and LORELEI Program No. HR0011-15-C-0115. The views and conclusions contained herein are those of the authors and should not be interpreted as necessarily representing the official policies, either expressed or implied, of DARPA, or the U.S. Government. The U.S. Government is authorized to reproduce and distribute reprints for governmental purposes notwithstanding any copyright annotation therein.

%% file: appendix/appendix.tex
\newpage

\section{Hyperparameters}

\begin{table}[h]
\centering
\begin{tabular}{l|l}
\toprule
train batch size        & 4    \\
eval batch size         & 4    \\
learning rate           & 3e-5 \\
accumulate grad batches & 4    \\
training epoches        & 6    \\
warmup steps            & 0    \\
weight decay            & 0    \\
\# gpus                 & 1    \\ \bottomrule
\end{tabular}
\caption{Hyperparameters for Training R-GQA.}
\end{table}

\section{Distinct Event Types in WikiEvent Ontology (as Compared to ACE05)}
\label{sec:distinct}

\begin{table}[h]
\centering
\resizebox{\columnwidth}{!}{
\begin{tabular}{l|l|l}
\toprule
Hierachy L1 & Hierachy L2 & Hierachy L3 \\ \midrule
ArtifactExistence & DamageDestroyDisableDismantle & Damage         \\
ArtifactExistence & DamageDestroyDisableDismantle & Destroy        \\
ArtifactExistence & DamageDestroyDisableDismantle & DisableDefuse  \\
ArtifactExistence & DamageDestroyDisableDismantle & Dismantle      \\
ArtifactExistence & DamageDestroyDisableDismantle & Unspecified    \\
ArtifactExistence & ManufactureAssemble           & Unspecified    \\
Cognitive         & IdentifyCategorize            & Unspecified    \\
Cognitive         & Inspection                    & SensoryObserve \\
Cognitive         & Research                      & Unspecified    \\
Cognitive         & TeachingTrainingLearning      & Unspecified    \\
Disaster          & DiseaseOutbreak               & Unspecified    \\
Disaster          & FireExplosion                 & Unspecified    \\
GenericCrime      & GenericCrime                  & GenericCrime   \\
Justice           & InvestigateCrime              & Unspecified    \\
Life              & Consume                       & Unspecified    \\
Life              & Illness                       & Unspecified    \\
Life              & Infect                        & Unspecified    \\
Medical           & Diagnosis                     & Unspecified    \\
Medical           & Intervention                  & Unspecified    \\
Medical           & Vaccinate                     & Unspecified    \\
Movement          & Transportation                & PreventPassage \\
Transaction       & Donation                      & Unspecified  \\ \bottomrule
\end{tabular}
}
\end{table}

\onecolumn
\input{appendix/analysis}

\newpage
\onecolumn
\section{Distribution Distances for Each Event Type (ACE05)}
\label{sec:distribution_distance_event_type}

\begin{table*}[h]
\centering
\begin{tabular}{l|cccc}
\toprule
Event Type              & Random   & AL & Context   & JointEnc   \\ \midrule
Movement:Transport             & 0.38   & 1.21  & 0.37    & 0.28     \\
Personnel:Elect                & 0.38   & 0.75  & 0.18    & 0.26     \\
Personnel:Start-Position       & 0.34   & 0.78  & 0.39    & 0.46     \\
Personnel:Nominate             & 0.43   & 0.52  & 0.49    & 0.31     \\
Personnel:End-Position         & 0.66   & 0.99  & 0.23    & 0.34     \\
Conflict:Attack                & 0.30   & 0.16  & 0.34    & 0.23     \\
Contact:Meet                   & 0.40   & 0.20  & 0.43    & 0.40     \\
Life:Marry                     & 0.54   & 0.32  & 0.25    & 0.24     \\
Transaction:Transfer-Money     & 0.38   & 0.38  & 0.41    & 0.42     \\
Conflict:Demonstrate           & 0.26   & 0.53  & 0.43    & 0.37     \\
Business:End-Org               & 0.67   & 0.25  & 0.64    & 0.28     \\
Justice:Sue                    & 0.63   & 1.09  & 0.47    & 0.48     \\
Life:Injure                    & 0.37   & 0.46  & 0.47    & 0.32     \\
Life:Die                       & 0.32   & 0.94  & 0.34    & 0.22     \\
Justice:Arrest-Jail            & 0.42   & 0.29  & 0.45    & 0.46     \\
Contact:Phone-Write            & 0.24   & 0.31  & 0.33    & 0.23     \\
Transaction:Transfer-Ownership & 0.24   & 0.32  & 0.30    & 0.22     \\
Business:Start-Org             & 0.78   & 0.86  & 0.45    & 0.30     \\
Justice:Execute                & 0.72   & 0.32  & 0.81    & 0.32     \\
Justice:Trial-Hearing          & 0.20   & 0.38  & 0.46    & 0.28     \\
Life:Be-Born                   & 0.77   & 0.31  & 0.41    & 0.28     \\
Justice:Charge-Indict          & 0.27   & 0.68  & 0.44    & 0.27     \\
Justice:Convict                & 0.47   & 0.55  & 0.49    & 0.48     \\
Justice:Sentence               & 0.13   & 0.41  & 0.34    & 0.57     \\
Business:Declare-Bankruptcy    & 0.27   & 0.84  & 0.37    & 0.30     \\
Justice:Release-Parole         & 0.38   & 0.22  & 0.46    & 0.46     \\
Justice:Fine                   & 0.42   & 0.22  & 0.43    & 0.41     \\
Justice:Pardon                 & 0.41   & 0.45  & 0.43    & 0.48     \\
Justice:Appeal                 & 0.62   & 0.35  & 0.31    & 0.63     \\
Justice:Extradite              & 0.37   & 0.83  & 0.55    & 0.56     \\
Life:Divorce                   & 0.32   & 1.01  & 0.30    & 0.20     \\
Business:Merge-Org             & 0.60   & 0.47  & 0.73    & 0.42     \\
Justice:Acquit                 & 0.59   & 0.71  & 0.49    & 0.57     \\ \midrule
Sum                            & 14.65  & 19.36 & 14.39   & 12.31    \\
Average                        & 0.43   & 0.55  & 0.42    & 0.36     \\ \bottomrule
\end{tabular}
\end{table*}

\newpage

\section{Generated Questions for Argument Roles in WikiEvent Ontology}

\input{tables/questions_1}

\input{tables/questions_2}

\input{tables/questions_3}

\input{tables/questions_4}

\input{tables/questions_5}


%% file: appendix/analysis.tex

\section{Further Findings and (Error) Analysis}

\paragraph{Error Cases and Remaining Challenges}

We conduct an analysis on the error cases and summarize representative causes and provide examples below:
\begin{itemize}[leftmargin=*]
    \item Lack of contextual understanding. For example, ``Earlier documents in the case have included embarrassing details about perks $[Welch]_{\sc Person}$ received as part of his \textbf{retirement} package from GE ..''. The model predicts the pronoun ``his'' which is closer to the trigger word as the final {\sc Person} argument for the retiring event, ignoring the better option ``Welch'' which is more informative. Also with the document-level contextual knowledge of the person ``Welch'' that appears frequently, it would be easier for the model to decide.
    \item Complex language usage such as idioms and metaphors (e.g. for the event with ``swept out of power'' as the trigger, the arguments' recall is very low).
    Addressing these phenomena is difficult since it requires richer knowledge about the background/culture. Plus, the special tokenization process further (e.g. BPE: Byte-Pair Encoding) further hurts the performance of extracting certain words that rarely appear.
    \item Inherent imperfectness of the datasets. The inter-annotator agreement for ACE05/WikiEvent is limited (under 85\%), so theoretically there is an upper bound for human performance as well. For example, we see that the head noun match (HM) score is strictly higher than the exact match (EM) in Section 4, and the gap mitigates as the performance gets higher (over 70\% F1). This demonstrates there is an ambiguity in determining the argument's boundary. Moreover, for the example in the first bullet point, predicting pronoun does not get credit -- while in a certain amount of training data it's permitted.
\end{itemize}

\paragraph{Influence of Similarity-based Retrieval}

\begin{figure}[h]
\centering
\resizebox{0.6\columnwidth}{!}{
\includegraphics{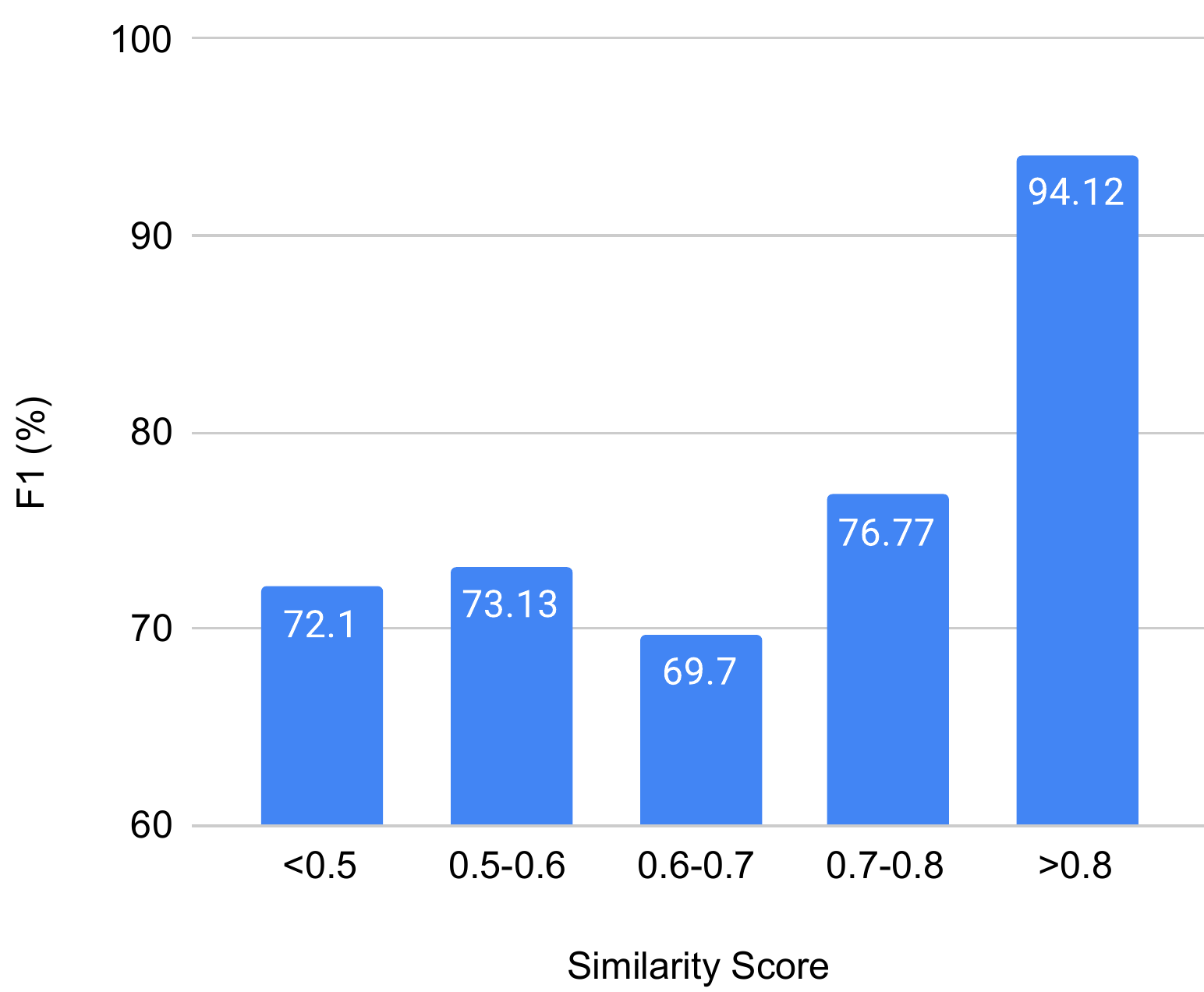}
}
\caption{R-GQA's performances on subsets of dataset (as the similarity scores grow).}
\label{fig:similarity_f1}
\end{figure}

In Figure~\ref{fig:similarity_f1}, we provide insights on how the similarity between the demonstration and current context affects the model's performance. We divide the original test set into five subsets, corresponding to the example's similarity score.
It is observed there is a trend that when the similarity score grows, performance of the model also grows, especially when the similarity is over 0.7. This to a certain extent shows the benefits of augmenting the current context with a more similar demonstration as the prompt.

%% file: tables/questions_1.tex
\begin{table*}[h]
\resizebox{\textwidth}{!}{
\begin{tabular}{llll}
\toprule
{\bf Event Type} & {\bf Argument Role} & {\bf Question} \\ \midrule
ArtifactExistence.DamageDestroyDisableDismantle.Damage        & Damager               & who is the damaging agent?                        \\
                                                              & Artifact              & what is being damaged?                            \\
                                                              & Instrument            & what is the instrument used in the damage?        \\
                                                              & Place                 & where the damage takes place?                     \\
ArtifactExistence.DamageDestroyDisableDismantle.Destroy       & Destroyer             & who is the destroying agent?                      \\
                                                              & Artifact              & what is being destroyed?                          \\
                                                              & Instrument            & what is the instrument used in the destroy?       \\
                                                              & Place                 & where the destroy takes place?                    \\
ArtifactExistence.DamageDestroyDisableDismantle.DisableDefuse & Disabler              & who is the disable agent?                         \\
                                                              & Artifact              & what is being disabled?                           \\
                                                              & Instrument            & what is the instrument used in the disable?       \\
                                                              & Place                 & where the disable takes place?                    \\
ArtifactExistence.DamageDestroyDisableDismantle.Dismantle     & Dismantler            & who is the dismantle agent?                       \\
                                                              & Artifact              & what is being dismantled?                         \\
                                                              & Instrument            & what is the instrument used in the dismantle?     \\
                                                              & Components            & who is being dismantled?                          \\
                                                              & Place                 & where the dismantle takes place?                  \\
ArtifactExistence.DamageDestroyDisableDismantle.Unspecified   & DamagerDestroyer      & who is the damaging agent?                        \\
                                                              & Artifact              & what is being destroyed                           \\
                                                              & Instrument            & what is the instrument used in the destroy        \\
                                                              & Place                 & where the destroy takes place?                    \\
ArtifactExistence.ManufactureAssemble.Unspecified             & ManufacturerAssembler & what is the manufacutring agent?                  \\
                                                              & Artifact              & what is being manufactured?                       \\
                                                              & Components            & what is the components used for the manufacture?  \\
                                                              & Instrument            & what is the instrument used in the manufacture?   \\
                                                              & Place                 & where the manufacutring takes place?              \\
Business:Declare-Bankruptcy                                   & Org                   & What declare bankruptcy?                          \\
                                                              & Place                 & Where the merger takes place?                     \\
Business:End-Org                                              & Org                   & What is ended?                                    \\
                                                              & Place                 & Where the event takes place?                      \\
Business:Merge-Org                                            & Org                   & What is merged?                                   \\
Business:Start-Org                                            & Agent                 & Who is the founder?                               \\
                                                              & Org                   & What is started?                                  \\
                                                              & Place                 & Where the event takes place?                      \\
Cognitive.IdentifyCategorize.Unspecified                      & Identifier            & who is the identifier?                            \\
                                                              & IdentifiedObject      & what is being identified?                         \\
                                                              & IdentifiedRole        & what is being identified as?                      \\
                                                              & Place                 & where the identifiying takes place?                 \\
Cognitive.Inspection.SensoryObserve                           & Observer              & who is the observer?                              \\
                                                              & ObservedEntity        & what is being observed?                           \\
                                                              & Instrument            & what is the instrument used in the observe?       \\
                                                              & Place                 & where the observe takes place?                    \\
Cognitive.Research.Unspecified                                & Researcher            & who is the researcher?                            \\
                                                              & Subject               & what is being researched?                         \\
                                                              & Means                 & what is being used for the research?              \\
                                                              & Place                 & where the research takes place?                   \\
Cognitive.TeachingTrainingLearning.Unspecified                & TeacherTrainer        & who is the teaching agent?                        \\
                                                              & FieldOfKnowledge      & what is being taught?                             \\
                                                              & Learner               & who is being taught?                              \\
                                                              & Means                 & what is being used for the teaching               \\
                                                              & Institution           & where is the teaching at institution              \\
                                                              & Place                 & where the teaching takes place?                   \\
Conflict.Attack.DetonateExplode                               & Attacker              & Who is the denotating agent?                      \\
                                                              & Target                & who is the target of the attack?                  \\
                                                              & Instrument            & What is the instrument used in the attack?        \\
                                                              & ExplosiveDevice       & what is the explosive device?                     \\
                                                              & Place                 & Where the detonation takes place?                 \\
                                                              
\bottomrule       
\end{tabular}
}
\end{table*}

%% file: tables/questions_2.tex
\begin{table*}[t]
\resizebox{\textwidth}{!}{
\begin{tabular}{llll}
\toprule
Conflict.Demonstrate.DemonstrateWithViolence & Demonstrator        & who is demonstrating agent?                                   \\
                                             & Regulator           & who is the regulator?                                         \\
                                             & VisualDisplay       & what is the visual display?                                   \\
                                             & Topic               & what is the topic for the demonstration?                      \\
                                             & Target              & who is the target of the demonstration?                       \\
                                             & Place               & where the demonstration takes place?                          \\
Conflict.Demonstrate.Unspecified             & Demonstrator        & who is demonstrating agent?                                   \\
                                             & Regulator           & who is the regulator?                                         \\
                                             & VisualDisplay       & what is the visual display?                                   \\
                                             & Topic               & what is the topic for the demonstration?                      \\
                                             & Target              & who is the target of the demonstration?                       \\
                                             & Place               & where the demonstration takes place?                          \\
Conflict:Attack                              & Attacker            & Who is the attacking agent?                                   \\
                                             & Instrument          & What is the instrument used in the attack?                    \\
                                             & Place               & Where the attack takes place?                                 \\
                                             & Target              & Who is the target of the attack?                              \\
                                             & Victim              & Who is the target of the attack?                              \\
Conflict:Demonstrate                         & Entity              & Who is demonstrating agent?                                   \\
                                             & Place               & Where the demonstration takes place?                          \\
Contact.Contact.Broadcast                    & Communicator        & who is communicating agents?                                  \\
                                             & Recipient           & who is the recipient?                                         \\
                                             & Instrument          & What is the instrument used in the communication?             \\
                                             & Topic               & what is the communicating topic?                              \\
                                             & Place               & Where it takes place?                                         \\
Contact.Contact.Correspondence               & Participant         & who is communicating agents?                                  \\
                                             & Instrument          & What is the instrument used in the communication?             \\
                                             & Topic               & what is the communicating topic?                              \\
                                             & Place               & Where it takes place?                                         \\
Contact.Contact.Meet                         & Participant         & Who are meeting?                                              \\
                                             & Topic               & what is the topic of the meeting                              \\
                                             & Place               & Where the meeting takes place?                                \\
Contact.Contact.Unspecified                  & Participant         & who is communicating agents?                                  \\
                                             & Topic               & what is the communicating topic?                              \\
                                             & Place               & Where it takes place?                                         \\
Contact.Prevarication.Unspecified            & Communicator        & who is communicating agents?                                  \\
                                             & Recipient           & who is communicating agents?                                  \\
                                             & Topic               & what is the communicating topic?                              \\
                                             & Place               & Where it takes place?                                         \\
Contact.RequestCommand.Unspecified           & Communicator        & who is communicating agents?                                  \\
                                             & Recipient           & who is communicating agents?                                  \\
                                             & Topic               & what is the communicating topic?                              \\
                                             & Place               & Where it takes place?                                         \\
Contact.ThreatenCoerce.Unspecified           & Communicator        & who is communicating agents?                                  \\
                                             & Recipient           & who is communicating agents?                                  \\
                                             & Topic               & what is the communicating topic?                              \\
                                             & Place               & Where it takes place?                                         \\
Contact:Meet                                 & Entity              & Who are meeting?                                              \\
                                             & Place               & Where the meeting takes place?                                \\
Contact:Phone-Write                          & Entity              & Who is communicating agents?                                  \\
                                             & Place               & Where it takes place?                                         \\
Control.ImpedeInterfereWith.Unspecified      & Impeder             & who is the impeder agent?                                     \\
                                             & ImpededEvent        & what is the impede event?                                     \\
                                             & Place               & where the impede takes place?                                 \\
Disaster.Crash.Unspecified                   & DriverPassenger     & Who is responsible for the transport event?                   \\
                                             & Vehicle             & What is the vehicle used to transport the person or artifact? \\
                                             & CrashObject         & what is being crashed into?                                   \\
                                             & Place               & where the transport takes place?                              \\
Disaster.DiseaseOutbreak.Unspecified         & Disease             & what broke out?                                               \\
                                             & Victim              & Who is the harmed person?                                     \\
                                             & Place               & Where the disease takes place?                                \\
Disaster.FireExplosion.Unspecified           & FireExplosionObject & what caught fire?                                             \\
                                             & Instrument          & What is the instrument used in the explosion?                 \\
                                             & Place               & where the explosion takes place?                              \\
GenericCrime.GenericCrime.GenericCrime       & Perpetrator         & who committed a crime?                                        \\
                                             & Victim              & Who is the target of the crime?                               \\
                                             & Place               & Where the crime takes place?                                  \\
 \bottomrule       
\end{tabular}
}
\end{table*}

%% file: tables/questions_3.tex
\begin{table*}[t]
\small
\resizebox{0.9\textwidth}{!}{
\begin{tabular}{llll}
\toprule
Justice.Acquit.Unspecified           & JudgeCourt        & What is the judge?                                         \\
                                     & Defendant         & Who is the defendant?                                      \\
                                     & Crime             & what is the crime being acquitted?                         \\
                                     & Place             & Where the acquit takes place?                              \\
Justice.ArrestJailDetain.Unspecified & Jailer            & Who is the arresting agent?                                \\
                                     & Detainee          & Who is jailed or arrested?                                 \\
                                     & Crime             & what is the crime being arrested?                          \\
                                     & Place             & Where the person is arrested?                              \\
Justice.ChargeIndict.Unspecified     & Prosecutor        & Indicated by whom?                                         \\
                                     & Defendant         & Who is indicted?                                           \\
                                     & JudgeCourt        & Who was the judge or court?                                \\
                                     & Crime             & what is the crime being charged?                           \\
                                     & Place             & Where the indictment takes place?                          \\
Justice.Convict.Unspecified          & JudgeCourt        & Who is the judge or court?                                 \\
                                     & Defendant         & Who is convicted?                                          \\
                                     & Crime             & what is the crime being convicted?                         \\
                                     & Place             & Where the conviction takes place?                          \\
Justice.InvestigateCrime.Unspecified & Investigator      & Who is the investigator?                                   \\
                                     & Defendant         & Who is investigated?                                       \\
                                     & Crime             & what is the crime being investigated?                      \\
                                     & Place             & Where the investigation takes place?                       \\
Justice.ReleaseParole.Unspecified    & JudgeCourt        & Who will release?                                          \\
                                     & Defendant         & Who is released?                                           \\
                                     & Crime             & what is the crime being released?                          \\
                                     & Place             & Where the release takes place?                             \\
Justice.Sentence.Unspecified         & JudgeCourt        & Who is the judge or court?                                 \\
                                     & Defendant         & Who is sentenced?                                          \\
                                     & Crime             & what is the crime being sentenced?                         \\
                                     & Sentence          & what is the sentence?                                      \\
                                     & Place             & Where the sentencing takes place?                          \\
Justice.TrialHearing.Unspecified     & Prosecutor        & Who is the prosecuting agent?                              \\
                                     & Defendant         & Who is on trial?                                           \\
                                     & JudgeCourt        & Who is the judge or court?                                 \\
                                     & Crime             & what is the crime being tried?                             \\
                                     & Place             & Where the trial takes place?                               \\
Justice:Acquit                       & Adjudicator       & Who was the judge or court?                                \\
                                     & Defendant         & Who was acquitted?                                         \\
Justice:Appeal                       & Adjudicator       & Who was the judge or court?                                \\
                                     & Place             & Where the appeal takes place?                              \\
                                     & Plaintiff         & What is the plaintiff?                                     \\
Justice:Arrest-Jail                  & Agent             & Who is the arresting agent?                                \\
                                     & Person            & Who is jailed or arrested?                                 \\
                                     & Place             & Where the person is arrested?                              \\
Justice:Charge-Indict                & Adjudicator       & Who was the judge or court?                                \\
                                     & Defendant         & Who is indicted?                                           \\
                                     & Place             & Where the indictment takes place?                          \\
                                     & Prosecutor        & Indicated by whom?                                         \\
Justice:Convict                      & Adjudicator       & Who is the judge or court?                                 \\
                                     & Defendant         & Who is convicted?                                          \\
                                     & Place             & Where the conviction takes place?                          \\
Justice:Execute                      & Agent             & Who carry out the execution?                               \\
                                     & Person            & Who was executed?                                          \\
                                     & Place             & Where the execution takes place?                           \\
Justice:Extradite                    & Agent             & Who is the extraditing agent?                              \\
                                     & Person            & Who is being extradited                                    \\
                                     & Destination       & Where the person is extradited to?                         \\
                                     & Origin            & Where is original location of the person being extradited? \\
Justice:Fine                                  & Adjudicator       & Who do the fining?                                            \\
                                              & Entity            & What was fined?                                               \\
                                              & Place             & Where the fining Event takes place?                           \\
Justice:Pardon                                & Adjudicator       & Who do the pardoning?                                         \\
                                              & Defendant         & Who was pardoned?                                             \\
                                              & Place             & Where the pardon takes place?                                 \\
Justice:Release-Parole                        & Entity            & Who will release?                                             \\
                                              & Person            & Who is released?                                              \\
                                              & Place             & Where the release takes place?                                \\
Justice:Sentence                              & Adjudicator       & Who is the judge or court?                                    \\
                                              & Defendant         & Who is sentenced?                                             \\
                                              & Place             & Where the sentencing takes place?                             \\
Justice:Sue                                   & Adjudicator       & Who is the judge or court?                                    \\
                                              & Defendant         & Who is sued against?                                          \\
                                              & Place             & Where the suit takes place?                                   \\
                                              & Plaintiff         & Who is the suing agent?                                       \\

 \bottomrule       
\end{tabular}
}
\end{table*}

%% file: tables/questions_4.tex
\begin{table*}[t]
\small
\resizebox{\textwidth}{!}{
\begin{tabular}{llll}
\toprule
Justice:Trial-Hearing                         & Adjudicator       & Who is the judge or court?                                    \\
                                              & Defendant         & Who is on trial?                                              \\
                                              & Place             & Where the trial takes place?                                  \\
                                              & Prosecutor        & Who is the prosecuting agent?                                 \\
Life.Consume.Unspecified                      & ConsumingEntity   & what is the consuming agent?                                  \\
                                              & ConsumedThing     & what is consumed?                                             \\
                                              & Place             & where the consuming takes place?                              \\
Life.Die.Unspecified                          & Victim            & Who died?                                                     \\
                                              & Place             & Where the death takes place?                                  \\
                                              & Killer            & Who is the attacking agent?                                   \\
                                              & MedicalIssue      & what is the medical issue                                     \\
Life.Illness.Unspecified                      & Victim            & who is victim?                                                \\
                                              & DeliberateInjurer & who is the deliberate injurer                                 \\
                                              & Disease           & what is the disease or sickness?                              \\
                                              & Place             & where the event takes place?                                  \\
Life.Infect.Unspecified                       & Victim            & who is victim?                                                \\
                                              & InfectingAgent    & who infected?                                                 \\
                                              & Source            & what is the infect from?                                      \\
                                              & Place             & where the event takes place?                                  \\
Life.Injure.Unspecified                       & Victim            & Who is the harmed person?                                     \\
                                              & Injurer           & Who is the attacking agent?                                   \\
                                              & Instrument        & What is the device used to inflict the harm?                  \\
                                              & BodyPart          & what is the body part being harmed?                           \\
                                              & MedicalCondition  & what is the medical issue?                                    \\
                                              & Place             & Where the injuring takes place?                               \\
Life:Be-Born                                  & Person            & Who is born?                                                  \\
                                              & Place             & Where the birth takes place?                                  \\
Life:Die                                      & Agent             & Who is the attacking agent?                                   \\
                                              & Instrument        & What is the device used to kill?                              \\
                                              & Place             & Where the death takes place?                                  \\
                                              & Victim            & Who died?                                                     \\
Life:Divorce                                  & Person            & Who are divorced?                                             \\
                                              & Place             & Where the divorce takes place?                                \\
Life:Injure                                   & Agent             & Who is the attacking agent?                                   \\
                                              & Instrument        & What is the device used to inflict the harm?                  \\
                                              & Place             & Where the injuring takes place?                               \\
                                              & Victim            & Who is the harmed person?                                     \\
Life:Marry                                    & Person            & Who are married?                                              \\
                                              & Place             & Where the marriage takes place?                               \\
Medical.Diagnosis.Unspecified                 & Treater           & who diagnosed the patient?                                    \\
                                              & Patient           & who is diagnosed?                                             \\
                                              & SymptomSign       & what is the symptom?                                          \\
                                              & MedicalCondition  & what is the medical condition?                                \\
                                              & Place             & where the event takes place?                                  \\
Medical.Intervention.Unspecified              & Treater           & what treated the patient?                                     \\
                                              & Patient           & who is treated?                                               \\
                                              & MedicalIssue      & what is the medical issue?                                    \\
                                              & Instrument        & What is the instrument used in the treatment?                 \\
                                              & Place             & Where the treatment takes place?                              \\
Medical.Vaccinate.Unspecified                 & Treater           & what treated the patient?                                     \\
                                              & Patient           & who is treated?                                               \\
                                              & VaccineTarget     & who is the target of the vaccination?                         \\
                                              & VaccineMethod     & what is the method of the vaccination?                        \\
                                              & Place             & Where the vaccination takes place?                            \\ 
Movement.Transportation.Evacuation            & Transporter       & Who is responsible for the transport event?                   \\
                                              & PassengerArtifact & Who is being transported?                                     \\
                                              & Vehicle           & What is the vehicle used to transport the person or artifact? \\
                                              & Origin            & Where the transporting originated?                            \\
                                              & Destination       & Where the transporting is directed?                           \\
Movement.Transportation.IllegalTransportation & Transporter       & Who is responsible for the transport event?                   \\
                                              & PassengerArtifact & Who is being transported?                                     \\
                                              & Vehicle           & What is the vehicle used to transport the person or artifact? \\
                                              & Origin            & Where the transporting originated?                            \\
                                              & Destination       & Where the transporting is directed?                           \\
Movement.Transportation.PreventPassage        & Transporter       & Who is responsible for the transport event?                   \\
                                              & PassengerArtifact & Who is being transported?                                     \\
                                              & Vehicle           & What is the vehicle used to transport the person or artifact? \\
                                              & Preventer         & who is preventing the transport?                              \\
                                              & Origin            & Where the transporting originated?                            \\
                                              & Destination       & Where the transporting is directed?                           \\
Movement.Transportation.Unspecified           & Transporter       & Who is responsible for the transport event?                   \\
                                              & PassengerArtifact & Who is being transported?                                     \\
                                              & Vehicle           & What is the vehicle used to transport the person or artifact? \\
                                              & Origin            & Where the transporting originated?                            \\
                                              & Destination       & Where the transporting is directed?                           \\
                                              
                                              \bottomrule       
\end{tabular}
}
\end{table*}

%% file: tables/questions_5.tex
\begin{table*}[t]
\small
\resizebox{\textwidth}{!}{
\begin{tabular}{llll}
\toprule

Movement:Transport                            & Agent             & Who is responsible for the transport event?                   \\
                                              & Artifact          & Who is being transported?                                     \\
                                              & Destination       & Where the transporting is directed?                           \\
                                              & Origin            & Where the transporting originated?                            \\
                                              & Vehicle           & What is the vehicle used to transport the person or artifact? \\
Personnel.EndPosition.Unspecified             & Employee          & Who is the employee?                                          \\
                                              & PlaceOfEmployment & Who is the employer?                                          \\
                                              & Position          & what is the position?                                         \\
                                              & Place             & Where the employment relationship ends?                       \\
Personnel.StartPosition.Unspecified           & Employee          & Who is the employee?                                          \\
                                              & PlaceOfEmployment & Who is the employer?                                          \\
                                              & Position          & what is the position?                                         \\
                                              & Place             & Where the employment relationship begins?                     \\
Personnel:Elect                               & Entity            & Who voted?                                                    \\
                                              & Person            & Who was elected?                                              \\
                                              & Place             & Where the election takes place?                               \\
Personnel:End-Position                        & Entity            & Who is the employer?                                          \\
                                              & Person            & Who is the employee?                                          \\
                                              & Place             & Where the employment relationship ends?                       \\
Personnel:Nominate                            & Agent             & Who is the nominating agent?                                  \\
                                              & Person            & Who are nominated?                                            \\
Personnel:Start-Position                      & Entity            & Who is the employer?                                          \\
                                              & Person            & Who is the employee?                                          \\
                                              & Place             & Where the employment relationship begins?                     \\
Transaction.Donation.Unspecified              & Giver             & Who is the donating agent?                                    \\
                                              & Recipient         & Who is the recipient?                                         \\
                                              & Beneficiary       & Who benefits from the transfer?                               \\
                                              & ArtifactMoney     & what is being donated?                                        \\
                                              & Place             & Where the transaction takes place?                            \\
Transaction.ExchangeBuySell.Unspecified       & Giver             & Who is the selling agent?                                     \\
                                              & Recipient         & Who is the buying agent?                                      \\
                                              & AcquiredEntity    & Who was bought or sold?                                       \\
                                              & PaymentBarter     & how much was the selling?                                     \\
                                              & Beneficiary       & Who benefits from the transaction?                            \\
                                              & Place             & Where the sale takes place?                                   \\
Transaction:Transfer-Money                    & Beneficiary       & Who benefits from the transfer?                               \\
                                              & Giver             & Who is the donating agent?                                    \\
                                              & Place             & Where the transaction takes place?                            \\
                                              & Recipient         & Who is the recipient?                                         \\
Transaction:Transfer-Ownership                & Artifact          & Who was bought or sold?                                       \\
                                              & Beneficiary       & Who benefits from the transaction?                            \\
                                              & Buyer             & Who is the buying agent?                                      \\
                                              & Place             & Where the sale takes place?                                   \\
                                              & Seller            & Who is the selling agent? \\                      \bottomrule       
\end{tabular}
}
\end{table*}